# Optimal Sensor Placement in Body Surface Networks using Gaussian Processes

Emad Alenany

Systems Science and Industrial Engineering Department Binghamton University

4400 Vestal Pkwy E, Binghamton, NY 13902 ealenan1@binghamton.edu

Changqing Cheng

Systems Science and Industrial Engineering Department Binghamton University

4400 Vestal Pkwy E, Binghamton, NY 13902 ccheng@binghamton.edu

Abstract: This paper explores a new sequential selection framework for the optimal sensor placement (OSP) in Electrocardiography imaging networks (ECGI). The proposed methodology incorporates the use a recent experimental design method for the sequential selection of landmarkings on biological objects, namely, Gaussian process landmarking (GPLMK) for better exploration of the candidate sensors. The two experimental design methods work as a source of the training and the validation locations which is fitted using a spatiotemporal Gaussian process (STGP). The STGP is fitted using the training set to predict for the current validation set generated using GPLMK, and the sensor with the largest prediction absolute error is selected from the current validation set and added to the selected sensors. Next, a new validation set is generated and predicted using the current training set. The process continues until selecting a specific number of sensor locations. The study is conducted on a dataset of body surface potential mapping (BSPM) of 352 electrodes of four human subjects. A number of 30 sensor locations is selected using the proposed algorithm. The selected sensor locations achieved average $R^2 = 94.40\%$ for estimating the whole-body QRS segment. The proposed method adds to design efforts for a more clinically practical ECGI system by improving its wearability and reduce the design cost as well.

## I. Introduction

The leap forward in smart sensing, communication and cloud computation in the last decade brings to light the Internet of Things (IoT) technology, which has been constantly transforming the design and control of various complex engineering systems, from advanced manufacturing to smart energy and transportation [1]–[3]. Particularly, the IoT offers an unprecedented way to monitor health conditions and detect early onset of chronic diseases, including the cardiovascular disease (CVD). CVD is considered the most common cause of death globally, and around 31.1% of fatalities in 2016 were attributed to CVD worldwide [4]. Remarkably, optimal care management for CVD hinges on accurate monitoring of cardiac electrical activities, early detection of symptoms, and effective intervention.

Physiologically, the heart beating is regularized by the electrical impulse generated by the sinus node. As the heart undergoes depolarization and repolarization, the electrical current diffuses in the heart and cause subtle change of electrical potential on the body surface. As such, the cardiac electrophysiological activity can be monitored by an array of electrodes placed on the body surface via electrocardiogram (ECG). Notwithstanding the ease to operate and wide availability in clinical setting, the 12-lead ECG purveys merely low-resolution depiction of cardiac electrical potential, not suited to diagnosis for a host of cardiac ailments [5]. By contrast, the ECG imaging (ECGI) technique has prominent edges over the traditional 12-lead ECG. The ECGI system consists of a body sensor network (BSN) with a large number of electrodes and provides spatiotemporal surface potential mapping (BSPM) [6]. When coupled with the torso geometry obtained from CT scan, BSPM offers an alternative solution for faithful retrieval of electrophysiological pattern on the epicardium in a noninvasive manner. This further allows for the reconstruction of local electrophysiological events, identifying risky patients of cardiac arrhythmias and premature cardiac death, and epicardial or myocardial activation reconstruction [7]. An example of a BSN of 352 sensors, generated via map3d software [8], is shown in Figure 1.

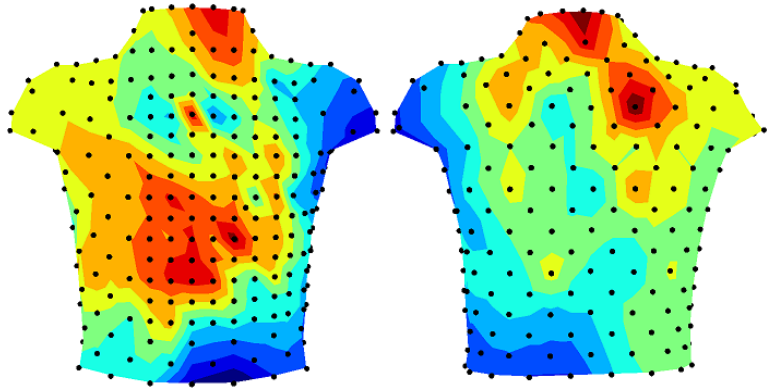

Figure 1 Body surface network of 352 electrodes: front torso (Left) and back torso (Right)

Yet, the large number of electrodes, that require steady skin contact, stymies the wide adoption of ECGI in healthcare practice[7]. Optimal selection of a small subset of sensors without affecting the data quality is conducive to reducing energy consumption to run the network [9] and enhance the wrearability. In addition, the optimal subset reduces the amount of data for storage and processing.

On the other hand, the optimal subset selection entails effective modeling of the spatiotemporal BSPM data, which is a confounding quest [10]. Such analysis is based on the methods matured mainly in geostatistics for spatial analysis which is referred to as Kriging (Gaussian process regression) [10]. A sensor network represents a complex system, whose behavior includes irregular and nonstationary patterns [2], [11], [12].

Gaussian process (GP) has been widely adopted in sensor placement problem for observing spatial events without considering spatiotemporal dynamics [13]. In contrast, the ECGI systems generate spatiotemporal data. The extension of GP to the spatiotemporal case is straightforward. In that case, the use of the GP model is named spatiotemporal GP (STGP). This research provides an investigation to model the ECGI spatiotemporal data with STGP using a smaller set of sensors in an ECGI system. A number of works have been done to tackle the optimal design of ECGI systems [14]–[17]. However, the current research contributes to ECGI system design by investigating the combination of STGP with one recent experimental design method for sensor selection, namely, Gaussian process landmarking (GPLMK) which will be described in the methodology section. This research presents a novel architecture for sequential sensor placement in BSN using a design of experiments-based method under a Gaussian process framework. This is done with the objective to better explore the candidate set of locations considered for inclusion in the optimal BSN design.

The optimal design of such sensors system represents a significant problem as it relates to the overall sensor network cost and its ability to achieve its stated objective(s) of the network [11]. Selecting the best locations for a subset of sensors could be sought so most of the information with the full number of sensors in the original design still gained. This is known as the sensor placement problem and is known to be NP-hard, and usually solved using greedy algorithms due to the submodularity property [13]. Submodularity property refers to that the expected gain in performance $E$ from adding a sensor $s$ to a small set of sensors $A$ is larger than the benefit of adding a sensor $s$ to a larger set of sensors $B$, i.e., $E(s \cup A) \geq E(s \cup B)$ [13].

The optimal sensor placement is known to be NP hard. This paper presents a novel sequential sensor placement method that incorporate a new experimental design method for better exploration of the solution space within the sequential selection framework based on STGP. The Gaussian process landmarking (GPLMK) approach recently developed in [18] uses GP framework and consider geometry in manifolds for locations selection in biological objects. The algorithm used in the current study is similar to [19]. However, the current study adds to the literature of the optimal sensor placement by exploring a novel sequential selection framework and STGP prediction and utilizing a newly developed experimental design method. The paper goes as follows, section II presents a short review of related literature, while section III presents the paper methodology. Then, section IV presents results and discussions and finally, section V presents conclusions and future work.

## II.   Literature review

This section provides a review of the related work to sequential sensing in BSN (lead selection) and a number of the related literature to the current research. Lead selection problem refers to the selection of a subset of the BSPM leads / locations / sites to obtain the most correct appraisal of the total signals from locations with no sensors selected or the selection of the best leads with the most informative extracted features for diagnosing specific diseases. The two categories are referred to as the most "signal" and the most "diagnostic" information sensor locations, respectively [20]. The current study considers the problem of optimal sensor placement for the getting the most "signal" information. Donnely et al. presents a summary of the leading work done in the area of optimal design of ECGI for around three decades [20]. They reassured the importance of retrospective analysis of BSPM spatial distributions for the optimal sensor placement in BSNs. In the following, a review about a number of applications of lead selection or optimal sensor placement (OBS) in BSN is provided.

Barr et al. presented one of the first studies that addressed studying the optimal lead selection for BSPM [15]. They used a dataset of 150 leads for 45 subjects and they aimed to estimate the QRS potential. Their algorithm works by finding what they called generator matrix $G(i,t)$ and coefficient matrix $A(i,t)$ that transform the generator matrix into the estimation matrix of surface potential $W_f(i,t)$ for each lead $i$ and time $t$ for the original measurements $V_s(i,t)$ such that:

$$W_f(i,t) = \sum_{j=1}^{N_g} A(i,t)G_f(i,t) \qquad (1)$$

where $N_g$ is the set of selected locations, and $A$ and $G$ is obtained using principal component analysis of the matrix $V_s V_s^T$ so that the mean squared error between the actual and estimated surface potential is minimized for any number of generators used. An iterative method is performed where a different set of locations is selected. The used procedure found that 24 locations is required to achieve an adequate level of estimation accuracy of average relative mean-square error of less than 4 %. Later, Lux et al., presents an optimal lead selection algorithm based on information index that measures how much the signal at one location is correlated with that on the remaining sites while taking signal variation into account [16]. They presented a sequential selection procedure starts with selecting the location with the largest information index, then add this lead to the set of selected leads, then the information index is recalculated for the remaining locations. Again, the lead with the highest information index is selected, and the iterative process continues till stopping condition is met. The authors used the system noise as the lower limit or stopping condition. So, the iterative algorithm would continue adding more leads till reaching close to that error limit. The system noise in that study is estimated from the restrained T-P segment to be 0.020

$mV$ root mean square (RMS). The potential estimation for the remaining set of leads is calculated as follows:

$$T = K_{mm}^{-1} K'_{me} \qquad (2)$$

where $T$ is the transformation matrix, $k_{mm}$ is the covariance matrix of measured potentials and $k_{me}$ is the cross-covariance matrix between measured and estimated potentials. The authors achieved an average RMS of $0.032\ mV$ with 30 selected locations.

For relatively recent studies, Finlay et al. used a sequential selection approach with multiple linear regression to sequentially select the next sensor location [17]. The method is similar to sequential forward selection method used in the feature selection method of the wrapper approach as the forward and backward feature selection methods. The authors used and compared two different performance measures to guide the selection procedure, namely, spatial root mean square and correlation coefficient between the actual and estimated mapping data. The algorithm continues till the selected leads equal to 32, where the number is selected to allow a comparison with a previous study used a different algorithm[17]. The dataset used includes 192 BSPM leads distributed uniformly of a rectangle sized 16 by 12. The dataset includes 116 subjects where 59 of them are for normal subjects, and the 57 for subjects with evidence of past myocardial infarction. The dataset is divided into 87 subjects used for training and 29 used for evaluating the selected leads from the training set. For every subject in the dataset, the QRS segment in addition to each fifth frame in the STT segment is used. This results in 13157 total training map frames and 4381 testing map frames. The optimal sensor placement problem is known to be NP hard [18]. Zhu et al. presented a recent study for an efficient optimal sequential design of ECGI based on the submodularity property and lazy greedy algorithm [14]. They utilized the sequentially selected sensors to predict for all sensor locations using a weighted kernel linear regression. They achieved a convergence in performance, i.e., the mean absolute error (MAE), at around 30 selected sensors. The prediction utilizing the selected locations led to the successful reconstruction of widely used ECG systems, i.e., 12-lead ECG, VCG, in addition to BSPM. The authors achieved $R^2 > 97\%$ for BSPM in the P, QRS, and T segments in the ECG signal. Besides, 12-lead ECG and VCG were predicted with $R^2 > 97.71\%$ and $99.44\%$, respectively. ECGI requires regular contact with torso skin to maintain the full flow of measurements, which is generally difficult to achieve in practice. The design for such a stochastic sensor network is introduced in to facilitate modelling of the discontinuous flow of information due to the 'stochastic sensor-skin contact' [9]. The authors used a sparse-kernel weighted regression to model the spatial pattern supported by an optimal kernel placement algorithm. Then, a sparse particle filter is used for the reduced dimensions of the spatial model parameters and updated when new observations obtained at the next time instance.

In their review about limited lead selection literature, Donnely et al. have gave a number of remarks about previous literature [20]. First, the approach used in the first studies focus on classical statistical methods, while the late studies employed current data mining procedures. The statistical methods are mainly focusing on the use of statistical transformation and could be somewhat seen as complex compared to the more recent ones which are clearer in their methodology. However, the second group of studies may be viewed as not as statistically robust as the first group of studies [21]. Second, it is found that around 24-35 electrodes are enough for a reasonable estimation accuracy of all electrodes sites measurements. Also, it is found that, for a specific number of chosen leads, there exists different lead sets with very close estimation accuracy. We believe the current study falls in the category of studies that use data mining techniques with the inclusion of experimental design methods.

After reviewing the previous work of the optimal sensor placement in BSNs, a number of the most related studies to the current work will be presented in the following. The used sequential selection methodology in this paper utilizes the active learning approach for sequentially adding selected sensors. Active learning aims to learn a function accurately and quickly [22]. One recent example for active learning to learn intermolecular potential energy surfaces (PES) as compared to utilizing the latin hypercube design (LHD) design is given in [23]. The authors have defined a set of reference testing set in which the potential prediction of the training set is evaluated.

The next selected point is selected to be the one with the highest prediction variance or the point with the largest absolute error in the test set. In the third active learning method, two training sets initialized with a random point are used to predict the testing set, and the point that has the largest prediction absolute mean error is selected and added to both training sets. Che et al. present a novel study to spot the stability regions in time-delay dynamic systems under a sequential design framework which integrates high and low fidelity models [19]. For a better exploration of the parameter space, the authors used generated inputs using the Sliced LHD method in which the points in one slice is being evaluated through GP with the current identified critical points to select the points with the largest expected improvement in the performance criteria. Later, high-fidelity simulation is used for estimation of the boundary for the limit state. The algorithm continues till convergence of the assessed boundary.

Few studies have focused on modeling BSPM using STGP. To the best of our knowledge, Graßhoff et al., provides the first example to model BSPM data using STGP, however, they have not considered the optimal sensor placement problem [24]. The authors have presented a framework for GP scalability for kernels with a nonstationary phase that extends structured kernel interpolation (SKI) [24]. The method entails the use of warping functions mixtures to obtain an understandable structure that exists in source separation problems as in biomedical signal processing. The authors have tested the method with a number of large biomedical datasets. The method is named the method wrapSKI.

The current research contributes to previous works by proposing an optimal sensor placement in BSN based on a data mining approach under a Gaussian process framework. It utilizes a recent experimental design method called GPLMK in a sequential design algorithm for efficient exploration of the search space [18]. Also, this research aims to use the STGP for the regression problem, benefiting from the widespread use of Gaussian processes for optimal sensor placement in the spatial domain. STGP is a powerful tool for modelling spatiotemporal processes over a long period of time [25]. STGP is recently used effectively to model BSPM data [24]. To the best of our knowledge, STGP has not been investigated before to model the BSPM under the sensor selection framework for BSN. As GP is known to computationally expensive, we also suggesting utilizing a recent experimental design method called the GPLMK for generating candidate testing points as for better exploration of the input space and reduces the computational effort required for prediction when selecting the next sensor location. The GPLMK method is used for generating landmarking on a three-dimension manifold as an input generating method for the candidate sensor locations set.

### III.     Methodology

The optimal sensor placement problem consists of selecting the best set $k$ of sensors from a large subset of sensors $V$ to optimize some criteria. For example, considering minimizing the entropy of the unobserved locations conditioned on the selected locations. Then the best set of sensors is:

$$A^* = \arg \min_{A \subset V: |A|=k} H(X_{V \setminus A} | X_A) \tag{3}$$

where $H(X_{V \setminus A} | X_A)$ represents the entropy at the non-selected locations (validation locations) $V \setminus A$ given the set of selected locations $A$ [13]. Using the identity $H(X_{V \setminus A} | X_A) = H(X_V) - H(X_A)$, it could be seen that:

$$A^* = \arg\ \min_{A \subset V: |A|=k} H(X_{V \setminus A} | X_A) = \arg\ \max_{A \subset V: |A|=k} H(X_A) \tag{4}$$

The optimal sensor placement is known to be NP hard. So, this paper presents a method that incorporate a new experimental design methods for better exploration of the solution space within the sequential selection framework called the Gaussian process landmarking (GPLMK) approach recently developed in [18]. Two methods usually used for the experimental design of selecting a small subset from a large number of elements, namely, sequential design and one-shot designs [26]. The sequential design depends on selecting the next location based on some criteria, for example, the location with the largest predictive variance [22]. This research proposes a sequential sensor selection procedure for the selection of a small subset of sensors in a body surface network using STGP.

Roadmap of the proposed methodology is showcased in . The first step includes generating a large number of candidate sensor locations (solution space). This step uses a recent method developed to select sequential landmarkings on biological objects called GPLMK. The generated candidates will work as the store among we obtain a set of testing locations $A_s$ in each iteration of the methodology as explained in the following. The sequence where the locations is selected with the GPLMK correspond to the sequence of the testing sets $A_s$ used in progressing iterations in the methodology. The rationale for this is that the GPLMK is a sequential experimental design method. A large sequence of geometrical landmarks (chosen to be 800) is generated using the method in [18]. Each selected landmark corresponds to the point with the maximum uncertainty conditioned on previously selected landmarks under Gaussian process framework as explained in detail later in this section. The GPLMK method is recently proposed for the automatic selection of geometrical landmarkings to enhance representation of biological objects. The number of landmarkings chosen to be very large as the generated sequence is not necessarily unique. This sequence of locations generated using GPLMK is referred to as $L$. Then, , the first 10 locations are generated using the LHD method to initialize the selected set of locations $A$. Here, we have chosen to initialize $A$ with the LHD method similar to [19]. However, it would make more sense to initialize the set of selected locations for GPLMK with the first 10 locations from the list of generated landmarkings. This will be considered for future work. Then, the next slice or next 40 locations from $L$ are used as the testing set $A_s$. Prediction is done over the testing set using the spatiotemporal Gaussian process (STGP) which is described in the next section. The first five locations with the largest sum of absolute prediction error are selected to be added to the selected locations $A$. This procedure is repeated till the number of selected locations reaches 30 sensors. Then, the overall performance measure is calculated. The set of selected locations $A$ are used to predict the potential for all the validation locations $V \setminus A$.

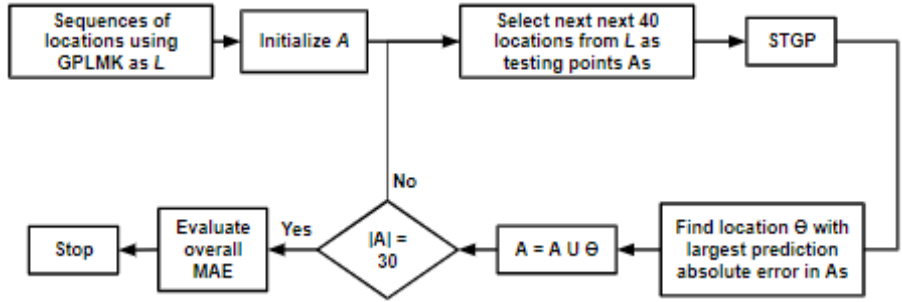

Figure 2 Methodology flowchart

A. GPLMK

GPLMK sequentially select landmarks on a Riemannian manifold by selecting the point with the largest prediction variance under the Gaussian process framework [18]. The algorithm input is the triangular mesh of the manifold of $T = (V, E)$ where $V$ is the set of vertices, $V \subset R^3$, and $E$ is the set of edges between the vertices.

The procedure starts with calculating the discrete heat kernel $K$ using the squared exponential function. Then, the Gaussian and mean curvature for each vertex $x_i$ in $V$ is calculated as $k(x) = k_1(x)k_2(x)$ and $\eta(x) = k_1(x) + k_2(x)$, respectively, where $k_1(x)$ and $k_2(x)$ are two principal curvatures at each vertex $x_i$. Then, the weight function at each vertex $x_i$ is calculated as follows:

$$w_{\lambda,\rho}(x_i) = \frac{\lambda |k(x_i)|^\rho}{\sum_{k=1}^{|V|} |k(x_k)|^\rho \, v(x_k)} + \frac{(1-\lambda)|\eta(x_i)|^\rho}{\sum_{k=1}^{|V|} |\eta(x_k)|^\rho \, v(x_k)} \forall x_i \in V \quad (5)$$

where $v(x_k)$ is the area of the Voronoi cell with a center at a vertex $x_i$, and $\lambda$ and $\rho$ are two parameters that control that ratio and fine-tune the energy surface of the Gaussian and mean curvatures, respectively. $W$ is the weight matrix with $w_{\lambda,\rho}(x_i)$ of each vertex are on its diagonal elements. To intensify for the effect of locations, reweighted heat kernel is calculated as:

$$k_t^{w_{\lambda,\rho}}(x_i, x_j) = \sum_{k=1}^{|V|} k_{t/2}(x_i, x_j) k_{t/2}(x_k, x_j) w_{\lambda,\rho}(x_k) v(x_k) =: (\widetilde{K}^T W \widetilde{K})_{ij}, \qquad (6)$$

where $k_{t/2}$ is an unweighted exponential kernel with bandwidth parameter $t/2$. The algorithm then selects the landmarking at step $(n+1)$ with the largest uncertainty score $\Sigma_{(n+1)}$ on $V$ conditioned on the current selected $n$ landmarks, $\xi_1, \dots, \xi_n$, where:

$$\Sigma_{(n+1)}(x_i) = k_t^{w_{\lambda,\rho}}(x_i, x_i) - k_t^{w_{\lambda,\rho}}(x_i, \xi_n^1) k_t^{w_{\lambda,\rho}}(\xi_n^1, \xi_n^1) k_t^{w_{\lambda,\rho}}(x_i, \xi_n^1) \qquad (7)$$

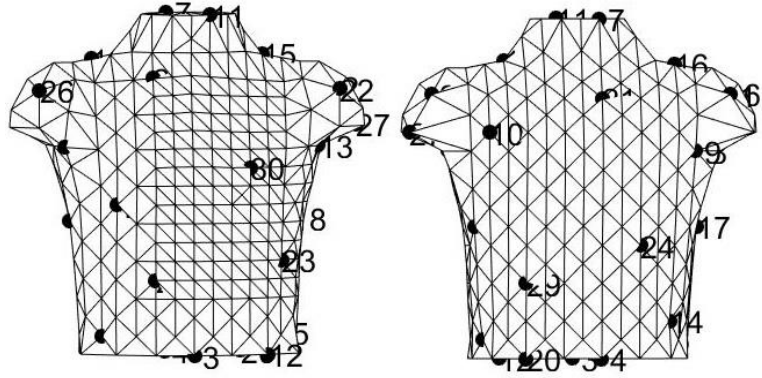

Figure 2 Body torso with 30 landmarks generated by GPLMK with the number next to each landmark represents the order of point. Front torso (Left) and Back torso (right).

Figure 2 shows the first 30 sequentially picked points using the GPLMK algorithm on the body torso with 352 vertices and 700 edges with $\lambda = \frac{1}{2}$ and $\rho = 1$.

B. Gaussian Process

Gaussian process (GP) is a non-parametric Bayesian approach that is used for regression and classification problems [27]. GP is completely defined by its mean and covariance function: $f(x) \sim GP(m(x), k(x, x'))$. Such that $m(x) = E[f(x)]$ is the expected value of the function at input $x$ and $k(x, x')$ represents the covariance function which determines the similarity between two input values x and $x'$. The covariance function is known also as the kernel such that: $cov(f(x), f(x')) = k(x, x') = E[(f(x) - m(x))(f(x') - m(x'))]$. The expressive capability of GP is mainly determined by its kernel (covariance function). The squared exponential (SE) is widely used to model smooth functions. It is also known as the radial basis function kernel (RBF) and has only two parameters the length scale $l$ and signal variance $\sigma$ per each input dimension, $x \in R^D$:

$$k(x, x') = \sigma^2 \exp\left(-\frac{|x - x'|^2}{2l^2}\right) \qquad (8)$$

The prediction of new test values $f_*$ (corresponding to new test inputs $X*$) given a set of training data $(X, Y)$ is done through conditioning over the known observations which have Gaussian distribution as well. The predictive mean and covariance of the posterior distribution are given by $f_*|X, Y, X_* \sim \mathcal{N}(\overline{f_*}, cov(\overline{f_*}))$, where

$$\overline{f_*} \triangleq E[f_*|X, Y, X_*] = K(X_*, X)[K(X, X) + \sigma_n^2 I]^{(-1)} Y, \qquad (9)$$

$$cov(\overline{f_*}) = K(X_*, X_*) - K(X_*, X)[K(X, X) + \sigma_n^2 I]^{-1} K(X, X_*) \qquad (10)$$

Hyperparameters optimization for kernel and mean functions usually done through maximization of the marginal (log) likelihood [31]:

$$\log p(Y|X) = -\frac{1}{2}y^T(K+\sigma_n^2 I)^{-1}Y - \frac{1}{2}\log|K+\sigma_n^2 I| - \frac{n}{2}\log 2\pi \tag{11}$$

Every observation ECGI is indexed by $(x, y, z, t)$, where $(x, y, z)$ are the coordinates of the sensor on the body torso and $t$ is the time instant. According to the American Heart Association recommendation (AHA), the $x, y,$ and $z$ axes refer to the transverse, longitudinal, and sagittal axes, respectively. The positive direction of the $x, y,$ and $z$ axes is defined to be left, inferior, and posterior, respectively. For x-axis, left means from the point of view of observer, as traditionally defined in physics. While the y and z axes positive directions is directed down, and back, respectively [28].

The covariance function between any two pairs of observations is defined as:

$$k\big((x,y,z,t),(x',y',z',t')\big) = k_x(x,x')k_y(y,y')k_z(z,z')k_t(t,t') \tag{12}$$

where $k_x, k_y,$ and $k_z$ are selected as the sum of rational quadratic with automatic relevance determination (ARD) and Matern-5/2 kernels for each space dimension and $k_t$ is a sum of a spectral mixture kernel (SM) with 12 components selected through experimentation, a white noise, and a non-stationary linear covariance. A similar covariance function was used in recent applications [29]–[31]. The spectral mixture kernel is defined as [32]:

$$k_t(t-t') = \sum_{q=1}^{Q} w_q\, e^{-2\pi^2(t-t')^2 v_q} \cos\big(2\pi(t-t')\mu_q\big) \tag{13}$$

The SM is derived by modelling the spectral density or Fourier transform by a mixture of $Q$ Gaussian, where $w_q, 1/\mu_q,$ and $1/\sqrt{v_q}$ represents the weights for each mixture component, the component periods, and the length-scales of each component [32].

The complexity of GP is known to be $O(n^3)$ for learning and $O(n^2)$ for storage. Such that n is the number of observations and $D$ is the data dimensions. For the BSN considered, $D = 4$ dimensions, $(x, y, z)$ coordinates for each location and time t for each observation.

Different approximation methods has been proposed to improve GP scalability, which could be categorized mainly into global and local approximation methods [33]. Of those, an approximation method named SKI has been proposed that utilize grid structure through Kronecker methods, and can reduce the complexity of GP to $O\left(Dn^{\frac{D+1}{D}}\right)$ for inference and learning and $O\left(Dn^{\frac{2}{D}}\right)$ for storage [34]. This method has been used previously with STGP studies due to the special structure in the spatial and time dimensions [29]–[31].

wrapSKI is a recent extension to SKI by using mixtures of non-stationary warping functions [35]. This allow SKI method suits for data with non-stationary phase. The method entails the use of warping functions mixtures to obtain an understandable structure that exists in source separation problems as in biomedical signal processing. The authors have provided the first example of modeling BSPM using STGP with a dataset of 62 leads to predict for 6 randomly chosen leads. The dataset used in the current study is of 4 human subjects with each of 1 beat length only, which makes wrapSKI is not suitable to apply and makes the SKI a better choice.

The dataset used in this study has 95, 56, and 119 unique values for $x, y,$ and $z$ dimensions, respectively. In addition, the temporal domain is of 1000 msec length. When using the SKI method, the spatial grid was reduced to around 10% of its size for computational tractability. We have realized that this hurt the prediction accuracy significantly. We have also performed multiple experiments that consider only a part of the time horizon when performing STGP calculations to reduce the required computations.

After exploring a reasonable number of GP approximation methods, we came across the laGP package [36].

We have found laGP is capable of handling larger data than other approximate GP methods we investigated. laGP refers to local approximate Gaussian Process. On the contrary to SKI, laGP is a local approximate to GP. It works by considering small dense matrices rather than sparse matrices. laGP approximates the prediction for a specific point $s$ through a part of the data $D_n(s) \subseteq D_N$ where $X_n(s)$ is the design matrix which mainly consists of the $n$ nearest neighbors to $s$. laGP starts use a greedy heuristic for selecting the design matrix $X_n(s)$. The procedure starts with a small number of neighbors to the point $s$ as $D_{n_0}(s)$, where $n_0 < n$. The next points will be added sequentially under one of simple objectives close to mean-square prediction error. The procedure is repeated until selecting $n$ observations in $D_n(s)$. One used criteria is to select the next selected point $s_{j+1}$ and add to $X_j(s)$ to form $D_{j+1}(s)$ to minimize the empirical Bayes mean-square prediction error (MSPE)

$$J(s_{j+1}, s) = \mathbb{E}\{[Y(s) - \mu_{j+1}(s|D_{j+1}, \hat{\theta}_{j+1})]^2 | D_j(s)\} \tag{14}$$

Using proper approximations, the whole procedure including the GP calculations as in equations 7 and 8 can be done in $O(n^3)$. As $n \ll N$, then laGP will be much faster than the exact GP calculations with reasonable accuracy by carrying the advantages of working local [37].

In the current study, we are using SKI while selecting the optimal sensor locations using our methodology. After completing the optimal sensor locations, we divert to laGP for the calculations of the final performance measures.

## IV. Results and Discussions

We adopt a dataset from Physionet which contains the potential for a BSN with 352 sensors for 4 human subjects with myocardial infarction for around 1000 milliseconds [38]. In this study, we have used the first human subject in the dataset in the process of the optimal sensor placement using the proposed methodology. Then, we evaluated the prediction accuracy of the set of optimal selections for the locations with no sensor selected for each of the 4 human subjects.

The $R^2$ and mean absolute error ($MAE$) are used to evaluate the accuracy of potential prediction in the validation locations using the selected sensors. $R^2$ and $MAE$ are calculated for the data of the validation locations only. Equations **Error! Reference source not found.** and **Error! Reference source not found.** show the calculations for $R^2$ and MAE, where $f(s_i, t)$ represent the measured potential at a specific validation location $s_i$ at time $t_i$ and $f^*(s_i, t)$ represents the predicted potential for the same location at the same time instance. N represents the number of validation points; those not included in modeling training and equals 322 and T is the total number of time points and equal to 1000 millisecond.

$$R^2 = 1 - \frac{T}{N} \cdot \frac{[f(s_i, t) - f^*(s_i, t)]^2}{[f(s_i, t) - \bar{f}^*(s_i, t)]} \tag{15}$$

$$MAE = \frac{1}{NT} \cdot \sum_{i=1}^{N} \sum_{t=1}^{T} |f(s_i, t) - f^*(s_i, t)| \tag{16}$$

Figure 3 shows the optimal sensor placement based on the proposed methodology. A number of 22 and 8 locations are selected on the front and back torso, respectively. Most of the sensors appear to bundle around the heart position on the front torso which agrees with the result in [14].

Figure 4 shows the MAE for the overall prediction of the validation locations using the sequential selection algorithm. With the optimal selection of the proposed methodology, the $R^2 = 86.11\%, MAE = 0.012$ with 30 selected sensors.

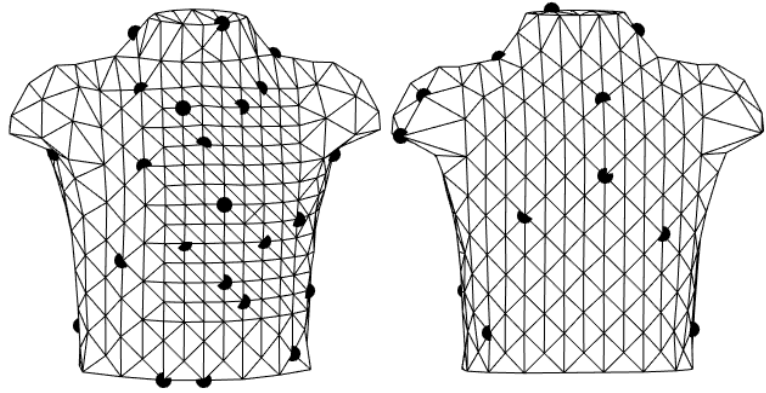

Figure 3 Optimal sensor placement

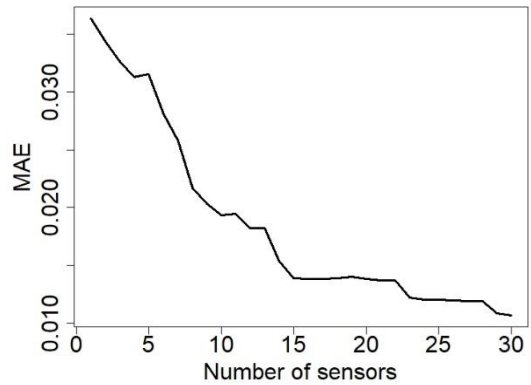

Figure 4 $MAE$ with number sensors

With experimentation, we evaluate the accuracy of the selected optimal sensors to predict the QRS segment of the ECG signal to reduce the computational requirements and also to obtain a reasonable estimation accuracy. Figure 8 shows a typical ECG signal with the QRS segment shown in red. The QRS is considered a crucial segment of the ECG signal and has been considered mainly to test the performance of the optimal sensor placement in BSN [15]–[17]. The QRS segment corresponds to the depolarization of the ventricular myocardium, and is useful to infer about events related to ventricular conduction abnormalities [7]. A recent study shows the significant information that QRS in BSPM can hold to accurately classify the Arrhythmogenic Right Ventricular Cardiomyopathy (ARVC) patients [39].

To study the potential of the optimal sensors to predict the QRS segment within the ECG signal of the validation set, we subset the data corresponding to the QRS segment for all the sensors locations, and use those of the optimal locations in Figure 5 as training data and apply laGP to predict for QRS for the validation set.

Table 1 shows the prediction accuracy for the QRS for 4 human subjects using the 30 locations from the sensor selection with the proposed methodology, in addition to a uniformly placed set of sensors. The used methodology achieved average $R^2 = 94.40\%$, while the uniform selection with the same number of sensors with $R^2 = 82.75\%$. This shows the potential of the used methodology to select a parsimonious set of sensor locations on the body torso to estimate the total-body QRS BSPM with acceptable accuracy. It is shown that the uniform selection shows high variability in $R^2$ measure. This could be attributed, for example, to the natural differences in the heart size in different human subjects. Table 1 also shows the performance of estimating the whole ECG signal for the validation locations using laGP. It can be shown that the performance with our methodology outperforms the uniform selection. However, the $R^2$ is reduced significantly when utilizing the whole ECG signal for the training set.

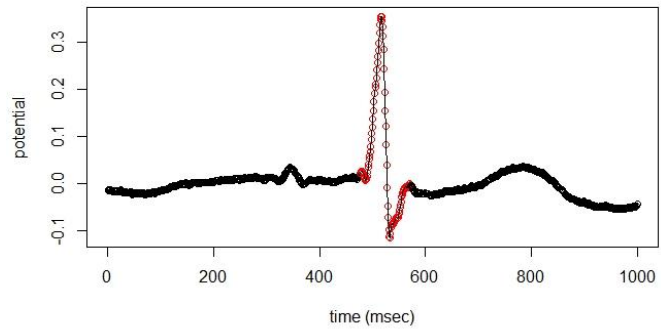

Figure 8 QRS segment (in red) within the ECG signal from one location of patient 3 in the dataset

Table 1 Performance of optimal and uniform selection for the QRS and the whole ECG signal

|  | GPLMK | | Uniform | |
| --- | --- | --- | --- | --- |
| Segment | MAE | R2 | MAE | R2 |
| QRS | 0.038 ± 0.009 | 94.40 ± 2.53 | 0.054 ± 0.007 | 82.75 ± 10.75 |
| Full ECG signal | 0.012 | 86.11 | 0.01 | 79.15 |

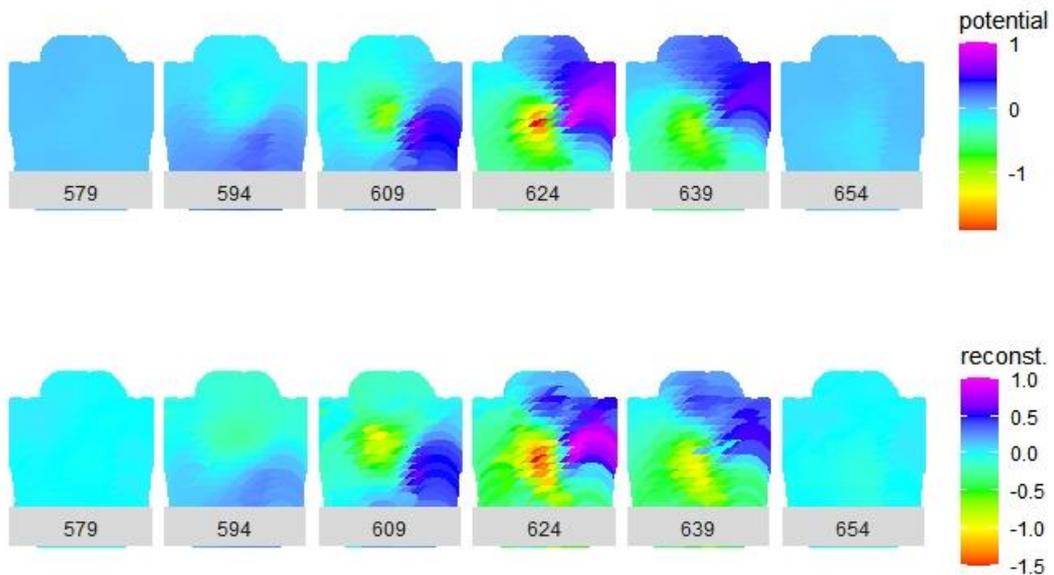

Figure 9 Sample of original (top) vs. reconstructed (down) QRS frames for the front torso

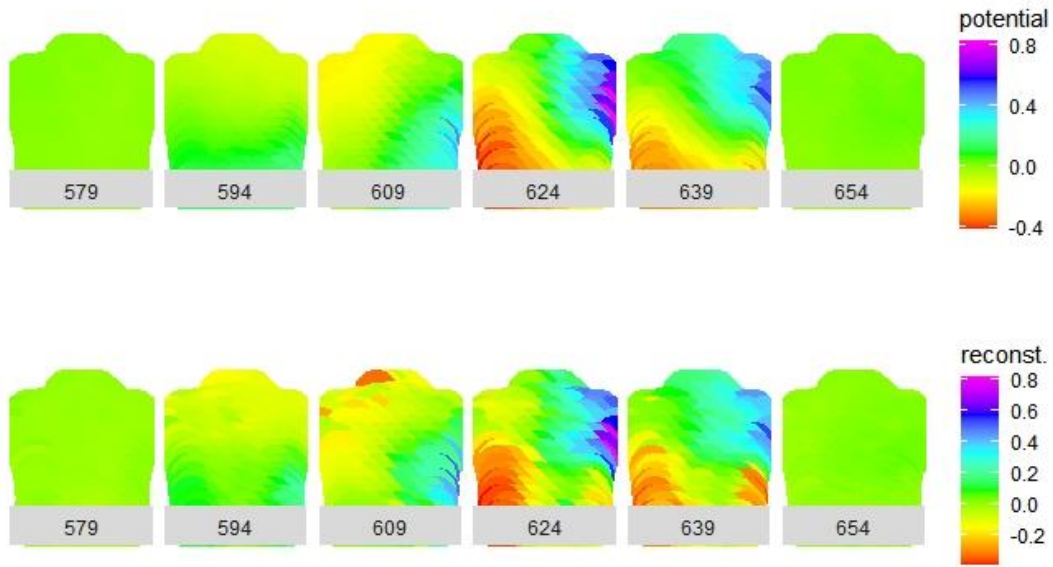

Figure 10 Sample of original (top) vs. reconstructed (down) QRS frames for the back torso

Figures 9 and 10 shows a sample of the QRS frames (starting with the QRS onset and following frames with 15 msec intervals) for original and reconstructed BSPM for the front and back torso, respectively. This exemplify the capability of the optimal selected sensors to estimate the potential at the locations with no sensors. Although the number of selected locations in the posterior torso is lower than on the anterior torso, we could realize that this does not significantly affect the estimation as shown in figure 10.

**V.     Conclusions**

In this study, we present a methodology for the optimal sequential sensor placement in BSN using STGP. The study shows the potential of the proposed methodology of combining a recent experimental design method named GPLMK and Gaussian process within sequential selection framework. The current study adds to the literature of OSP in BSNs with the ultimate goal of having a more clinically practical BSN.

Future work should include other parts of the ECG signal other than the QRS. The used methodology needs to be studied with a longer time horizon and a larger dataset of larger number of human subjects. Also, future work would consider using the wapSKI method that is proposed to extend GP scalability for data with non-stationary phase as in biomedical signals within the sensor selection framework.